\documentclass[runningheads, orivec]{llncs}
\usepackage[T1]{fontenc}
\usepackage{graphicx}
\usepackage{amsfonts}
\usepackage{amsmath}
\usepackage{booktabs}

\usepackage{orcidlink}
\usepackage{marvosym}   

\begin{document}
\title{CharDiff-LP: A Diffusion Model with Character-Level Guidance for License Plate Image Restoration}
\titlerunning{CharDiff-LP}
%
\author{Kihyun Na\inst{1}\orcidlink{0009-0001-3827-5371} \and
Gyuhwan Park\inst{1}\orcidlink{0009-0008-7153-5113} \and
Injung Kim\inst{1}\orcidlink{0000-0003-4439-6097}\,\Letter}
\authorrunning{K. Na et al.}
%
\institute{Department of Computer Science and Electrical Engineering, General Graduate School, Handong Global University, Pohang 37554, Republic of Korea
\email{kevinna95@gmail.com, park01129@naver.com, ijkim@handong.edu}\\
\url{http://deeplearning.handong.edu}
}
\maketitle              
\begin{abstract}
License plate image restoration is important not only as a preprocessing step for license plate recognition but also for enhancing evidential value, improving visual clarity, and enabling broader reuse of license plate images.
Existing restoration methods either lack explicit textual guidance or rely on global string-level priors that overlook the spatial locality of individual characters, thereby limiting their effectiveness under severe degradation.
To address this, we propose a novel diffusion-based framework with character-level guidance, CharDiff-LP, which effectively restores and recognizes severely degraded license plate images captured under realistic conditions.
CharDiff-LP leverages fine-grained character-level priors extracted through an external segmentator and an Optical Character Recognition (OCR) module tailored for low-quality license plate images.
For precise and focused guidance, CharDiff-LP incorporates a novel \textit{Character-guided Attention through Region-wise Masking (CHARM)} module, which ensures that each character's guidance is restricted to its own region, thereby avoiding interference with other regions.
In experiments, CharDiff-LP significantly outperformed baseline restoration models in both restoration quality and recognition accuracy, achieving a 28.3\% relative reduction in character error rate (CER) on the Roboflow-LP dataset compared with the best-performing baseline.
\end{abstract}
\section{Introduction}
\begin{figure}[t]
\centering
\includegraphics[width=\textwidth]{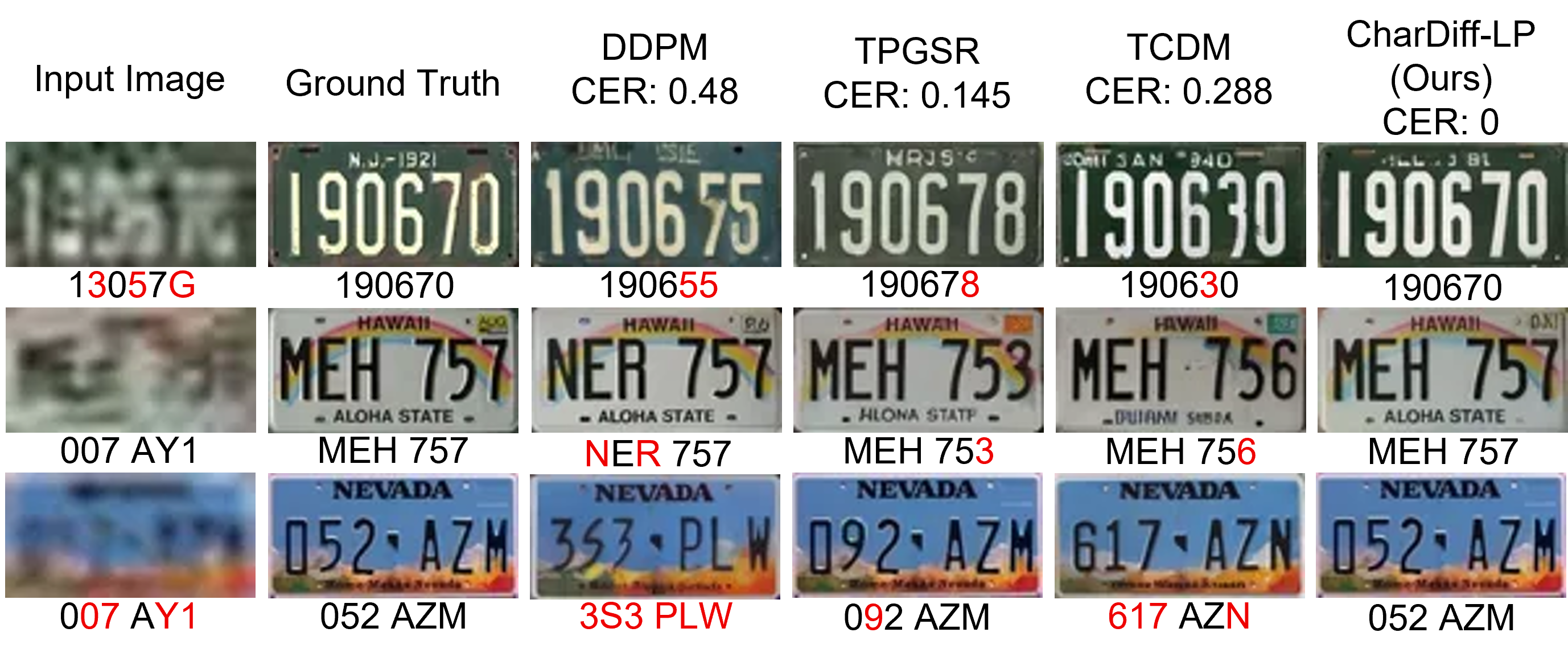}
\caption{
The results of diffusion-based image restoration models on severely degraded license plate images. Characters in red font indicate erroneously recognized and incorrectly restored characters (CER: character error rate).
}
\label{fig:visual_comparison}
\end{figure}

License Plate Recognition (LPR) is a core component of intelligent transportation systems (ITS), supporting applications such as traffic surveillance, automated toll collection, and access control. In practice, however, license plate images acquired in real-world environments often suffer from low resolution, motion blur, compression artifacts, and adverse lighting, which substantially deteriorate the performance of LPR systems~\cite{kaiser2021degradedlp1,liu2024irregularlp2}.

To alleviate this, previous work has explored image restoration techniques~\cite{Chao2015srcnn,zhang2018residual,wang2018esrgan,yang2020learning,zamir2022restormer,LI2022srdiff} as a preprocessing step for LPR. Beyond boosting recognition accuracy, license plate restoration is also important for verifying recognition results, increasing evidential value, and improving the visual clarity of interfaces such as advanced driver assistance systems (ADAS). GAN-based~\cite{liu17dpgan,lee18srgan,lin2021lprgan} and more recent diffusion-based restoration models~\cite{alhalawani24diffplate,gong25lpdiff} have shown substantial improvements. However, these methods do not explicitly exploit the textual information of license plates and, under severe degradation, often synthesize images that fail to preserve proper character shapes or even resemble different characters, as illustrated in Fig.~\ref{fig:visual_comparison}.

Several studies introduced text-guided restoration techniques into the LPR domain by designing loss functions that leverage OCR outputs during training~\cite{lee2020super,nascimento2023super,pan2024lpsrgan,nascimento24lacd}. Although such training-time guidance improves restoration performance to some extent, the model still lacks explicit inference-time guidance regarding character identities and locations, which limits its effectiveness in restoring fine details under severe degradation.

In the field of scene text image super-resolution (STISR), recent models employ conditional diffusion guided by textual information extracted from the input image~\cite{zhou2023rgdiff,noguchi2024tcdm,singh2024dcdm}.
Because they guide the denoising process via global cross-attention, these methods do not fully exploit the spatial information of individual characters. As a result, each character-specific prior may not focus exclusively on its corresponding region, causing interference across neighboring regions. Moreover, these STISR models are not tailored to the structured and highly constrained nature of license plates.

To address these limitations, we propose \textit{CharDiff-LP}, a diffusion-based framework with inference-time character-level guidance for restoring and recognizing low-quality license plate images. CharDiff-LP integrates fine-grained character-level priors extracted by three modules specialized for low-quality license plates: an external character segmentator, an OCR model robust to degraded inputs, and a character encoder that converts individual characters into embeddings. The segmentator leverages the structural format of license plates to identify character regions, the OCR model is trained on heavily degraded images, and the character encoder provides more fine-grained features than conventional string-level text encoders.

A key component of our framework is the \textit{Character-guided Attention through Region-wise Masking (CHARM)} module, which applies spatially masked cross-attention so that each character region attends only to its corresponding character embedding, preventing inter-character interference. To the best of our knowledge, this is the first diffusion-based license plate restoration model to incorporate character-level guidance via spatially masked cross-attention.

We validate our framework on two LPR datasets: one with paired high- and low-quality images generated via synthetic degradations, and another with real low-quality dashcam images without high-quality counterparts. In experiments, CharDiff-LP consistently improved both restoration quality and text recognition accuracy over recently developed baseline models. On the Roboflow-LP dataset, it reduced the CER from 11.3\% to 8.1\% compared to the second-best model, TCDM~\cite{noguchi2024tcdm}, corresponding to a 28.3\% relative error reduction. These results demonstrate that incorporating character-level priors with spatially guided attention is highly effective for restoring severely degraded images in practical LPR scenarios.

Our main contributions are summarized as follows:
\begin{itemize}
    \item \textbf{CharDiff-LP}: We propose a diffusion-based framework tailored for license plate restoration and recognition, which incorporates character-level priors to guide the restoration process.
    \item \textbf{CHARM}: We introduce the Character-guided Attention through Region-wise Masking (CHARM) module, which injects individual character embeddings into their corresponding regions via masked cross-attention, preventing interference between neighboring characters.
    \item \textbf{Comprehensive Evaluation}: We demonstrate consistent improvements over baselines on two datasets with both synthetic and real-world degradations.
\end{itemize}

\section{Related Work}
\subsection{License Plate Restoration and Recognition}

Traditional LPR systems typically follow a two-stage pipeline of license plate detection and character recognition, and operate directly on the captured image without explicit restoration~\cite{silva2018wpod,laroca2021efficient,moussa2022lptransf}. Under realistic conditions, however, degradations such as low resolution, motion blur, and sensor noise lead to large drops in accuracy~\cite{silva2021flexible,ke2023ultra,liu2024improving,luo2024reallpr,na2025mflpr2}.

To mitigate this, many studies add an image restoration stage before recognition. In general, unconditional restoration methods treat license plate images as generic images and learn a mapping from low- to high-quality images using CNN- or GAN-based restoration~\cite{yang2018cnn,vasek2018license,kim2024afa,liu17dpgan,lee18srgan,lin2021lprgan} or more recent diffusion-based restoration~\cite{alhalawani24diffplate,gong25lpdiff}. These methods improve visual quality but do not explicitly leverage textual content, and under severe degradation they may hallucinate semantically or structurally incorrect characters. Text-guided methods incorporate OCR feedback via semantic supervision or OCR-based losses and discriminators~\cite{zou2019semanticlpr,lee2020super,nascimento2023super,pan2024lpsrgan,nascimento24lacd}. However, they mainly use text priors as training-time regularizers and still lack explicit, localized guidance at inference time.

\subsection{Scene Text Image Super-Resolution}

STISR shares similar goals with license plate restoration: enhancing both visual quality and text legibility. TSRN~\cite{wang2020tsrn} introduced a text-specific architecture with sequential residual blocks and a boundary-aware loss, highlighting the gap between synthetic and real-world degradations. Subsequent works leveraged pre-trained recognizers to provide auxiliary OCR losses~\cite{chen2021tbsrn}, but guidance was still absent at inference time. TPGSR~\cite{ma2023tpgsr} first introduced inference-time guidance by fusing a text prior from a recognizer with image features, and TATT~\cite{ma2022tatt} replaced local fusion with Transformer-based cross-attention for global string-level alignment. More recently, diffusion-based STISR models such as TCDM~\cite{noguchi2024tcdm} and DCDM~\cite{singh2024dcdm} inject textual information directly into the denoising process, but rely on global cross-attention and do not explicitly constrain each character prior to its spatial region.

Motivated by these limitations, we extend text-guided diffusion models from STISR to the LPR domain with two key enhancements. First, we extract reliable character-level priors using an external segmentator and an OCR model specialized for degraded license plate images. Second, we introduce the CHARM module, which injects character embeddings via region-wise masked cross-attention so that each prior focuses on its corresponding character region.

\section{Method}
\subsection{Overview}
\begin{figure}[t]
\centering
\includegraphics[width=\textwidth]{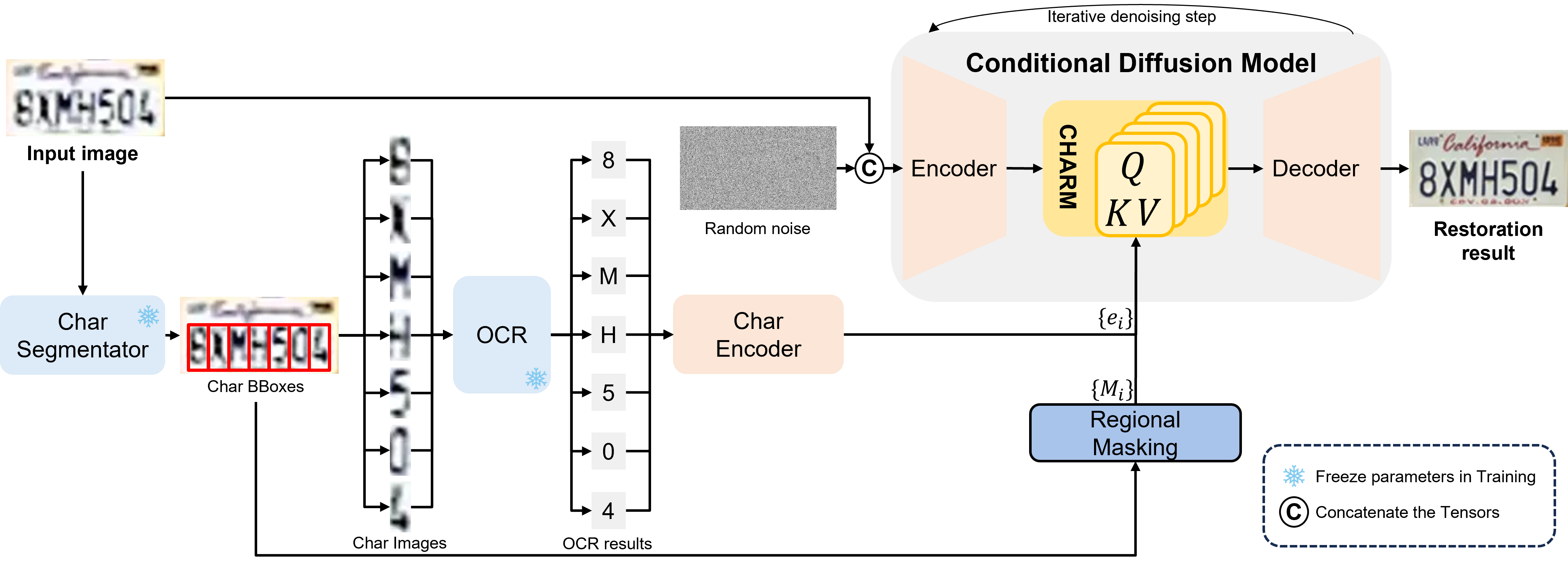} 
\caption{
The overview of the CharDiff-LP framework. Character-level priors—character embeddings and spatial masks—are extracted from the low-quality input image using an external segmentator and an OCR module. These priors are then injected into the diffusion model via the CHARM module to guide the restoration process at the character level throughout the denoising process.
}
\label{overview_CharDiff-LP}
\end{figure}

Figure~\ref{overview_CharDiff-LP} illustrates the overall process of the CharDiff-LP framework, which aims to restore a low-quality license plate (LP) image by incorporating fine-grained character-level guidance into the denoising process.

Given a low-quality input image $x^{LQ}$, CharDiff-LP segments individual characters with an external segmentator customized for license plate images. Then, a robust OCR model recognizes the segmented character images.

Subsequently, in addition to the low-quality input image, CharDiff-LP exploits two kinds of contextual priors: a sequence of recognized characters $\mathbf{y}=(y_1, ..., y_n)$, and their corresponding bounding boxes $\mathbf{B}=(B_1, ...,B_n)$. $y_i$'s are converted into the corresponding character embeddings $e_i$'s through the character encoder. The bounding boxes are converted into binary spatial masks $\mathbf{M}=(M_1, ..., M_n)$, $M_i \in \{0,1\}^{H \times W}$, where $H \times W$ denotes the feature map size. $M_i[u,v]$ is one for $(u,v)\in B_i$ and zero for other elements. The prior from each character consists of the character embedding extracted from the character class and the binary mask derived from its region, i.e., $c_i = (e_i, M_i)$.

The restoration process is based on a diffusion model employing a U-Net architecture. The character-level priors $(c_1, ..., c_n)$ are referenced through the CHARM module, which is placed in the middle of the U-Net. In the cross-attention mechanism of the CHARM module, the attention weight for referencing each character’s prior incorporates the spatial mask $M_i$, ensuring that the embedding $e_i$ from the $i$-th character guides only its corresponding region.

\subsection{Diffusion-based Restoration Model}
Our restoration model builds on DDPM~\cite{ho2020ddpm}, where a noise predictor $\epsilon_\theta(x_t', t, c)$ learns to denoise $x_t = \sqrt{\bar{\alpha}_t} x_0 + \sqrt{1 - \bar{\alpha}_t}\epsilon$ by minimizing:
\begin{equation}
\label{eq:ddpm_loss}
\mathcal{L}_{\text{cond}} = \mathbb{E}_{x_0, \epsilon, t} \left[ \left| \epsilon - \epsilon_\theta(x_t', t, c) \right|_2^2 \right].
\end{equation}
The conditioning information consists of a low-quality input image $x^{LQ}$ and textual priors $c$. The former is concatenated with $x_t$ to form $x_t' = \operatorname{Concat}(x_t, x^{LQ})$, while the latter is incorporated into the denoising process via cross-attention.

We compare three settings with respect to the text prior. 
The first setting, \textit{no text prior}, uses \( c = \emptyset \), where the model relies solely on \( x^{LQ} \), corresponding to the DDPM baseline. 
The second setting, \textit{string-level prior}, employs a single embedding for the entire string, \( c = e_{\mathbf{y}} \), similar to TCDM~\cite{noguchi2024tcdm}. 
The third setting, \textit{character-level prior (proposed)}, introduces a set of localized priors, 
\( c = \{(e_i, M_i)\}_{i=1}^n \), where each pair encodes the identity and spatial region of a character. Following TCDM, the text priors $c$ are injected into the middle block of the U-Net backbone in the second and third settings. The character-level priors are spatially localized via binary masks $M_i$, enabling more precise guidance for individual character regions.

\subsection{Character-Level Prior}
We construct character-level priors from the low-quality license plate image \( x^{LQ} \), which consist of semantic embeddings and spatial masks for individual characters, denoted as $\{(e_i, M_i)\}_{i=1}^n$. The extraction process comprises three stages: character segmentation, character recognition, and embedding projection.

\textbf{(1) Character Segmentation and Mask Generation}  
To localize character regions, we utilize a CRAFT-based text detector~\cite{baek2019craft} fine-tuned on an internal license plate dataset collected from~\cite{2025platesmania} that is independent of Roboflow-LP and Dashcam-LP. The detector outputs quadrilateral character regions, which are converted into axis-aligned bounding boxes $\{B_i\}$ through post-processing. For each bounding box, we generate a binary mask $M_i \in \{0,1\}^{H \times W}$, where only the elements inside the region are set to one. These masks are resized to match the resolution of the U-Net feature map used in the CHARM module. The segmentator is frozen during training.

\textbf{(2) Character Recognition}  
We crop the segmented character regions and feed them to the OCR module to predict their class labels. We adopt MGP-STR~\cite{wang2022mgpstr}, a strong scene-text recognizer, as the OCR module. To improve robustness to low-quality inputs, we further fine-tune it on artificially degraded license plate images, and keep the OCR model frozen during the training of the diffusion model.

\textbf{(3) Character Embedding}  
The recognized character labels are one-hot encoded, and then converted into character embeddings \( \{e_i\}_{i=1}^n \) by a lightweight Transformer encoder. Unlike the previous two modules, the embedding module is trained jointly with the diffusion model, allowing it to learn task-specific representations to guide the restoration process.

Consequently, CharDiff-LP extracts a set of localized priors that encode both the class and spatial location \( M_i \) of each character. These priors are delivered to the CHARM module to enable spatially focused and semantically informed image restoration.

\subsection{Character-guided Attention through Region-wise Masking (CHARM)}

\begin{figure}[t]
\centering
\includegraphics[width=0.64\textwidth]{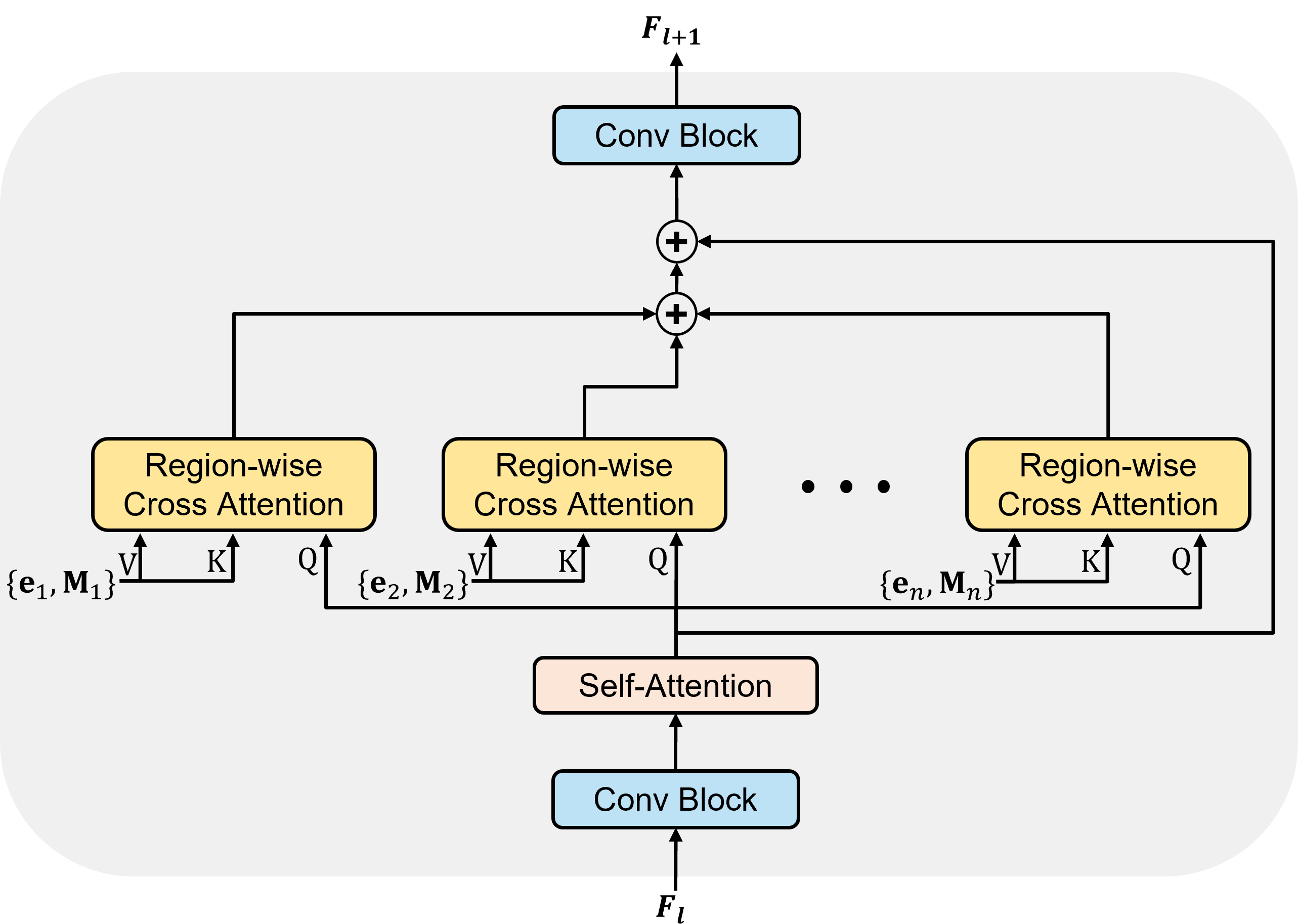}
\caption{
The detailed architecture of the CHARM module. Given a feature map $F_l$ from the U-Net encoder, the CHARM module injects the character embedding $e_i$ of each character only to the designated spatial region using the spatial mask $M_i$ via masked cross-attention.
}
\label{fig:charm_module}
\end{figure}

To inject character-level priors into the denoising process, we replace the standard string-level cross-attention in the U-Net middle block with the Character-guided Attention through Region-wise Masking (CHARM) module, a lightweight yet effective mechanism for spatially localized attention. The CHARM module combines the character embeddings $e_i$ with the intermediate feature map through masked cross-attention, as shown in Fig.~\ref{fig:charm_module}.

\noindent\textbf{Feature Preparation:}
Let \( F_l \in \mathbb{R}^{C \times H \times W} \) denote the input feature map of the CHARM module.
We first flatten the spatial dimensions and apply a linear projection to obtain the query matrix
\( Q \in \mathbb{R}^{C \times HW} \).
Each character embedding \( e_i \in \mathbb{R}^C \) is expanded to form key and value vectors
\( K_i, V_i \in \mathbb{R}^{C \times 1} \) via 1D convolution or linear layers.

\noindent\textbf{Region-masked Cross-Attention:}
For the \(i^{th}\) character, we compute attention weights between the query \( Q \) and the key \( K_i \),
applying a spatial mask \( M_i \in \{0,1\}^{H \times W} \), which is flattened to a vector in
\( \{0,1\}^{HW} \) during computation (we reuse the same symbol \(M_i\) by slight abuse of notation).
Let the channel dimension be \(d\) (i.e., \(Q \in \mathbb{R}^{d \times HW}\), \(K_i, V_i \in \mathbb{R}^{d \times 1}\)).
We first compute the masked attention as
\begin{equation}
    A_i = \text{softmax}\left( \frac{Q^\top K_i}{\sqrt{d}} + (1 - M_i)\cdot(-\infty) \right),
\end{equation}
where \(A_i \in \mathbb{R}^{HW}\).
Here, the term \( (1 - M_i)\cdot(-\infty) \) serves as a masking operation, effectively setting the attention scores of background regions to zero after the softmax operation.\footnote{In practice,
\(-\infty\) is implemented as a large negative constant.}
The attended feature \( h_i \in \mathbb{R}^{d \times HW} \) is then computed as
\begin{equation}
    h_i = V_i \cdot A_i^\top,
\end{equation}
and reshaped back to \( \mathbb{R}^{d \times H \times W} \).
Finally, the features from all characters are aggregated and added to the input feature map $F_l$ via a residual connection:
\begin{equation}
    F_{l+1} = F_l + \sum_{i=1}^n \text{Reshape}(h_i).
    \label{eq:eq5}
\end{equation}

\section{Experiments}
\subsection{Datasets}
We evaluate CharDiff-LP on two license plate datasets: Roboflow-LP and Dashcam-LP. Roboflow-LP contains 4,000 high-quality images from Roboflow Universe~\cite{roboflow-lp}; their low-quality counterparts were artificially generated using an extended Real-ESRGAN degradation pipeline~\cite{wang2021realesrgan} with additional perspective, contrast, and motion-blur augmentations to better mimic dashcam footage. The dataset is split into training, validation, and test sets with a 7:2:1 ratio.

Dashcam-LP comprises 10,000 images captured from vehicle dashcams under diverse driving conditions. Among them, 9,000 relatively clean images are used as high-quality training targets, and 1,000 challenging low-quality images (e.g., long distances, strong motion blur) are used as an unpaired real-world test set without corresponding high-quality references. For training, low-quality counterparts of the high-quality images are synthesized using the same degradation pipeline as in Roboflow-LP. We train separate models on the training splits of Roboflow-LP and Dashcam-LP and evaluate them on the corresponding test splits.

String-level annotations are available for all images, enabling recognition evaluation (CER and plate-level accuracy) even for unpaired real-world low-quality test images without corresponding high-quality references.

\subsection{Evaluation Metrics}
We evaluate both restoration quality and downstream recognition performance. On the Roboflow-LP dataset, where paired high-quality images are available, we measure PSNR, SSIM~\cite{wang2004ssim}, and LPIPS~\cite{zhang2018lpips} between the restored images and the ground-truth references. For license plate recognition (LPR), we compute the character error rate (CER) and plate-level accuracy based on exact string matching. Recognition performance on both Roboflow-LP and Dashcam-LP is evaluated by feeding the restored images into a fixed (frozen) MGP-STR recognizer; this recognizer is used solely for evaluation and is not jointly trained with any restoration model.

\subsection{Baseline Models}
We compared CharDiff-LP with three representative baseline models:  
(i) DDPM~\cite{ho2020ddpm}, a conditional diffusion model that takes only a low-quality image as input and serves as an unguided diffusion baseline;  
(ii) TPGSR~\cite{ma2023tpgsr}, a general STISR method that leverages textual priors extracted by a recognizer and incorporates them via local feature fusion; and  
(iii) TCDM~\cite{noguchi2024tcdm}, a diffusion-based STISR model that injects string-level textual priors into the denoising process through global cross-attention.  
For TPGSR and TCDM, we evaluated both the official pretrained models and variants fine-tuned on our license plate datasets. DDPM and CharDiff-LP were trained from scratch on each dataset.  
We also considered DiffPlate~\cite{alhalawani24diffplate} and LP-Diff~\cite{gong25lpdiff}, but excluded them from comparisons because they do not provide official implementations or adopt settings that are not directly comparable, such as multi-frame inputs.

\subsection{Implementation Details}
CharDiff-LP was implemented in PyTorch based on the DDPM framework~\cite{ho2020ddpm}. We used 1,000 diffusion steps with a linear noise schedule during training and 50 steps for inference. The noise predictor was optimized with the loss in Eq.~\ref{eq:ddpm_loss} using AdamW~\cite{loshchilov2017adamw} with a batch size of 64, a fixed learning rate of $3\times10^{-4}$, and 100k training steps. All input images were resized to $96\times 48$ pixels, reflecting the typical aspect ratio of U.S. license plates. For all methods, we followed the original implementation details when available and trained on a single NVIDIA RTX 4090 GPU under identical data splits and evaluation protocols.

\subsection{Comparison with Baseline Models}
\subsubsection{Quantitative Evaluation on Roboflow-LP}
\begin{table}[t]
\centering
\caption{
Quantitative evaluation results on the \textbf{Roboflow-LP} test set (paired real HQ and synthetic LQ images). CharDiff-LP achieves the best performance across all restoration and recognition metrics. Boldface indicates the best performance. (LPI: license plate images.)
}
\label{tab:quantitative-roboflow}
\begingroup
\setlength{\tabcolsep}{4pt}
\renewcommand{\arraystretch}{1.1}
\scriptsize
\begin{tabular}{lcccccc}
\toprule
\textbf{Method} & \textbf{Trained on LPI} & \textbf{PSNR}↑ & \textbf{SSIM}↑ & \textbf{LPIPS}↓ & \textbf{CER}↓ & \textbf{LPR Acc.}↑ \\
\midrule
DDPM (no text prior) & Yes & 23.98 & 0.8363 & 0.5022 & 0.212 & 0.514 \\
TPGSR (pretrained)   & No  & 18.85 & 0.6463 & 0.6221 & 0.425 & 0.458 \\
TPGSR (fine-tuned)           & Yes & 23.75 & 0.8216 & 0.4340 & 0.157 & 0.637 \\
TCDM (pretrained)    & No  & 18.91 & 0.6461 & 0.6124 & 0.404 & 0.476 \\
TCDM (fine-tuned)            & Yes & 24.16 & 0.8411 & 0.4114 & 0.113 & 0.667 \\
\textbf{CharDiff-LP (Ours)} & Yes & \textbf{24.35} & \textbf{0.8487} & \textbf{0.3832} & \textbf{0.081} & \textbf{0.697} \\
\bottomrule
\end{tabular}
\endgroup
\end{table}
\begin{table}[t]
\centering
\caption{
Quantitative evaluation results on the \textbf{Dashcam-LP} test set (unpaired real LQ images). CharDiff-LP achieves the best performance in both metrics. Boldface indicates the best performance. (LPI: license plate images.)
}
\label{tab:quantitative-dashcam}
\begingroup
\scriptsize
\setlength{\tabcolsep}{4pt}
\begin{tabular}{lccc}
\toprule
\textbf{Method} & \textbf{Trained on LPI} & \textbf{CER}↓ & \textbf{LPR Acc.}↑ \\
\midrule
DDPM (No text prior) & Yes & 0.2118 & 0.5458 \\
TPGSR (pretrained) & No  & 0.4473 & 0.2226 \\
TPGSR (fine-tuned)           & Yes & 0.1714 & 0.6266 \\
TCDM (pretrained)  & No  & 0.4339 & 0.2327 \\
TCDM (fine-tuned)            & Yes & 0.1613 & 0.6367 \\
\textbf{CharDiff-LP (Ours)} & Yes & \textbf{0.1512} & \textbf{0.6965} \\
\bottomrule
\end{tabular}
\endgroup
\end{table}
We evaluated CharDiff-LP using both restoration metrics (PSNR, SSIM~\cite{wang2004ssim}, and LPIPS~\cite{zhang2018lpips}) and recognition metrics (CER and LPR accuracy), and compared it against the baseline models. Table~\ref{tab:quantitative-roboflow} presents the evaluation results on the Roboflow-LP dataset, whose test set consists of pairs of real HQ and synthetic LQ images. The real HQ images were used as ground truth to compute the restoration metrics. `Trained on LPI' indicates that the model was trained or fine-tuned on the license plate image dataset, Roboflow-LP, starting from its pretrained weights. DDPM and CharDiff-LP were trained from scratch without any pretraining, using only LPI data.

In Table~\ref{tab:quantitative-roboflow}, the proposed model, CharDiff-LP, demonstrates superior performance across all metrics. Compared to the strongest baseline, TCDM (fine-tuned), CharDiff-LP improves PSNR by 0.19 dB, SSIM by 0.0076, and LPIPS by 0.0282. In addition, CharDiff-LP outperforms the baseline models in terms of recognition performance. Compared to the second-best model, TCDM (fine-tuned), CharDiff-LP reduces the Character Error Rate (CER) from 11.3\% to 8.1\%, corresponding to an absolute improvement of 3.2 percentage points and a relative error reduction of 28.3\%. It also achieves the highest plate-level LPR accuracy of 69.7\%. Given the poor quality of the test images (as shown in Fig.~\ref{fig:qualitative-roboflow}), these results demonstrate that the proposed character-level prior is highly effective in restoring global structure—crucial for identity preservation—as well as fine-grained details.

\subsubsection{Quantitative Evaluation on Dashcam-LP}
To assess generalization to authentic and uncontrolled degradations, we also evaluated recognition performance on the Dashcam-LP test set, which contains real low-quality images without high-quality counterparts (Table~\ref{tab:quantitative-dashcam}). `Trained on LPI' indicates trained or fine-tuned on the Dashcam-LP training split. Trained only on this dataset, CharDiff-LP still achieves the best performance, reducing CER from 16.1\% to 15.1\% compared to TCDM (fine-tuned) and improving plate-level LPR accuracy from 63.7\% to 69.7\% (a gain of 6.0 percentage points). These results confirm that the proposed character-level guidance is robust to challenging, unseen real-world degradation patterns.

\subsubsection{Qualitative Evaluation}
\begin{figure}[ht]
\centering
\includegraphics[width=\textwidth]{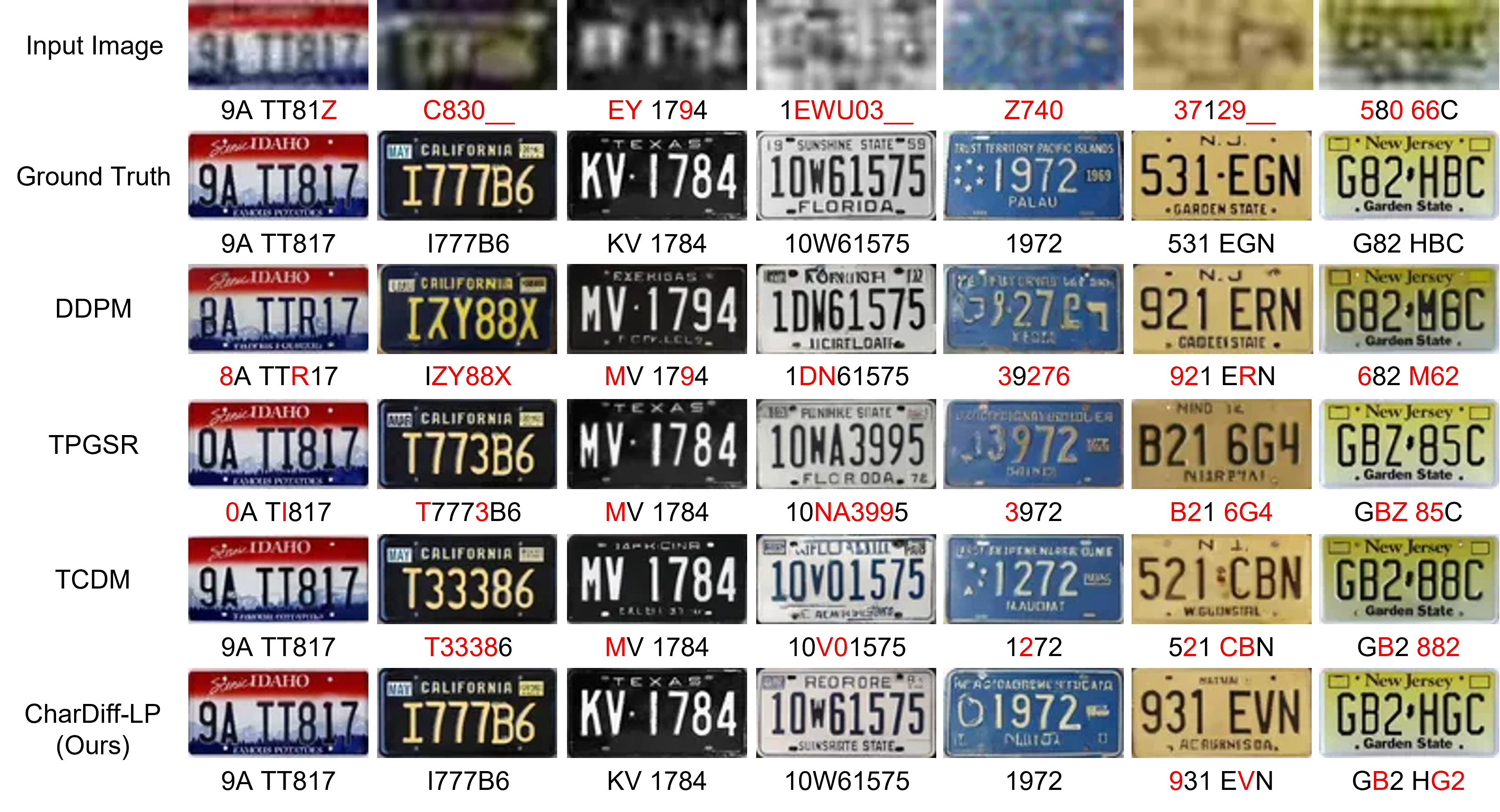}
\caption{
Restoration results on low-quality test images from the \textbf{Roboflow-LP} dataset. Red fonts indicate incorrectly restored characters and misrecognition results.
}
\label{fig:qualitative-roboflow}
\end{figure}
\begin{figure}[ht]
\centering
\includegraphics[width=\textwidth]{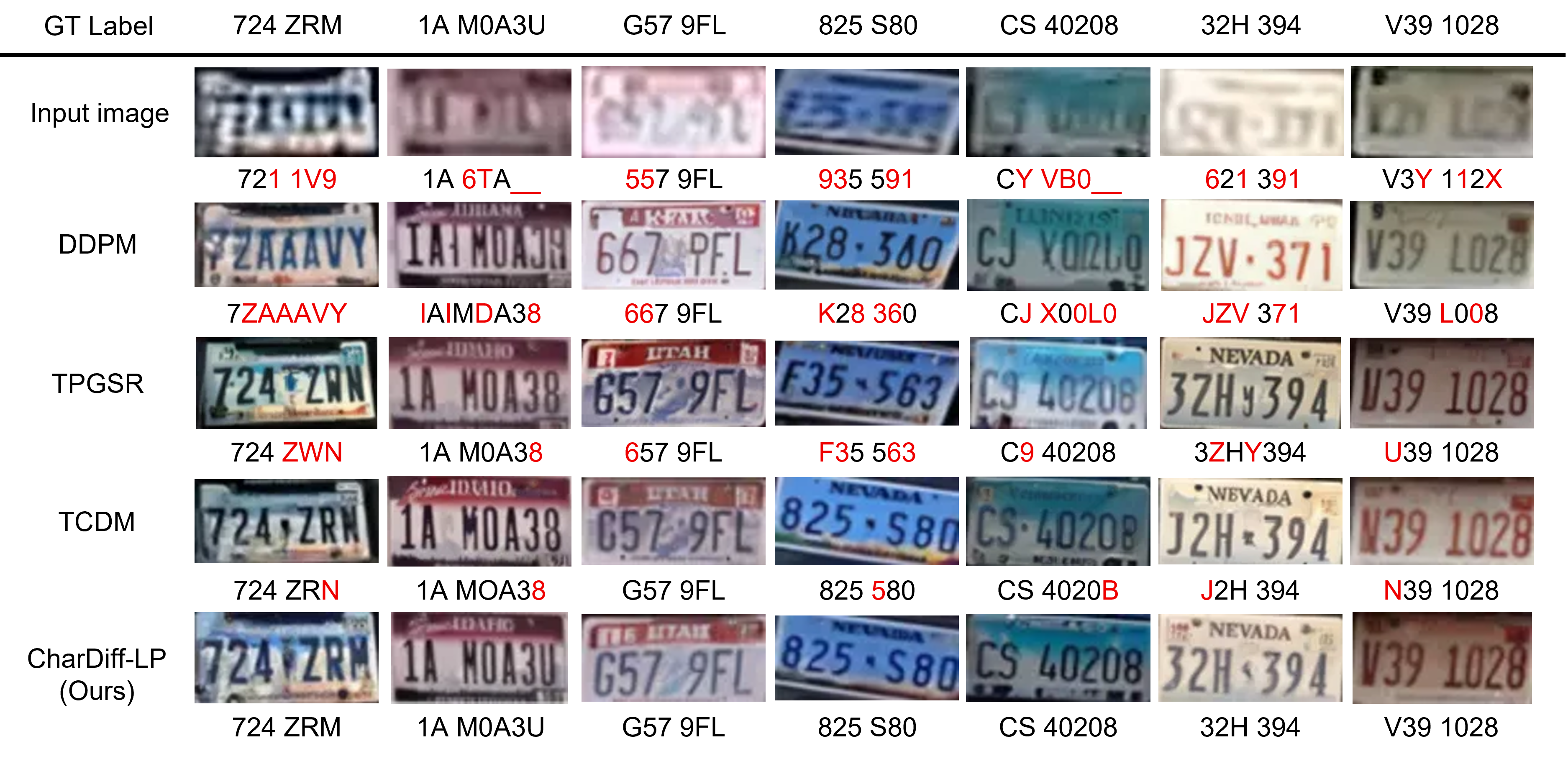}
\caption{
Restoration results of low-quality test images from the \textbf{Dashcam-LP} dataset. Red fonts indicate incorrectly restored characters and misrecognition results.
}
\label{fig:qualitative-dashcam}
\end{figure}

Fig.~\ref{fig:qualitative-roboflow} and~\ref{fig:qualitative-dashcam} illustrate the restoration results on samples from the Roboflow-LP and Dashcam-LP test sets. All methods enhance the perceptual quality of low-quality inputs, but the baseline models often produced artifacts, distorted character shapes, or identity changes (e.g., `I$\rightarrow$T', `7$\rightarrow$3', `B$\rightarrow$8' in Fig.~\ref{fig:qualitative-roboflow}). In contrast, CharDiff-LP generally yielded cleaner character shapes and better preserved the ground-truth identities across both synthetic and real-world degradations, which is consistent with the quantitative gains in CER and plate-level accuracy.
Failure cases predominantly arise when the input is severely degraded, rendering reliable character recovery infeasible for all methods.

\subsection{Ablation Studies}
\begin{table}[t]
\centering
\caption{
The results of ablation studies on the \textbf{Roboflow-LP} dataset (paired real HQ and synthetic LQ images), showing that each component incrementally improves restoration and recognition performance. The `DDPM + Character-level prior' model does not apply the CHARM module and references contextual information via global cross-attention.
}
\label{tab:ablation-roboflow-LP}
\begingroup
\setlength{\tabcolsep}{4pt}
\renewcommand{\arraystretch}{1.1}
\scriptsize
\begin{tabular}{lccccc}
\toprule
\textbf{Method Variant} & \textbf{PSNR}↑ & \textbf{SSIM}↑ & \textbf{LPIPS}↓ & \textbf{CER}↓ & \textbf{LPR Acc.}↑ \\
\midrule
DDPM (no text prior)           & 23.98 & 0.8363 & 0.5022 & 0.2118 & 0.514 \\
DDPM + String-level prior      & 24.21 & 0.8415 & 0.4478 & 0.1539 & 0.575 \\
DDPM + Character-level prior   & 24.30 & \textbf{0.8492} & 0.4210 & 0.1187 & 0.621 \\
CharDiff-LP (Ours)                & \textbf{24.35} & 0.8487 & \textbf{0.3832} & \textbf{0.0810} & \textbf{0.697} \\
\bottomrule
\end{tabular}
\endgroup
\end{table}

\begin{table}[t]
\centering
\caption{
The results of ablation studies on the \textbf{Dashcam-LP} dataset (unpaired real LQ images), showing that each component incrementally improves recognition performance.
}
\label{tab:ablation-dashcam-LP}
\setlength{\tabcolsep}{10pt}
{\normalsize
\begin{tabular}{lcc}
\toprule
\textbf{Method Variant} & \textbf{CER}↓ & \textbf{LPR Acc.}↑ \\
\midrule
DDPM (no text prior)& 0.2118 & 0.5458 \\
DDPM + String-level prior& 0.1709 & 0.6534 \\
DDPM + Character-level prior& 0.1614 & 0.6703 \\
CharDiff-LP (Ours)& \textbf{0.1512} & \textbf{0.6965} \\
\bottomrule
\end{tabular}
}
\end{table}

\begin{figure}[ht!]
\centering
\includegraphics[width=0.65\textwidth]{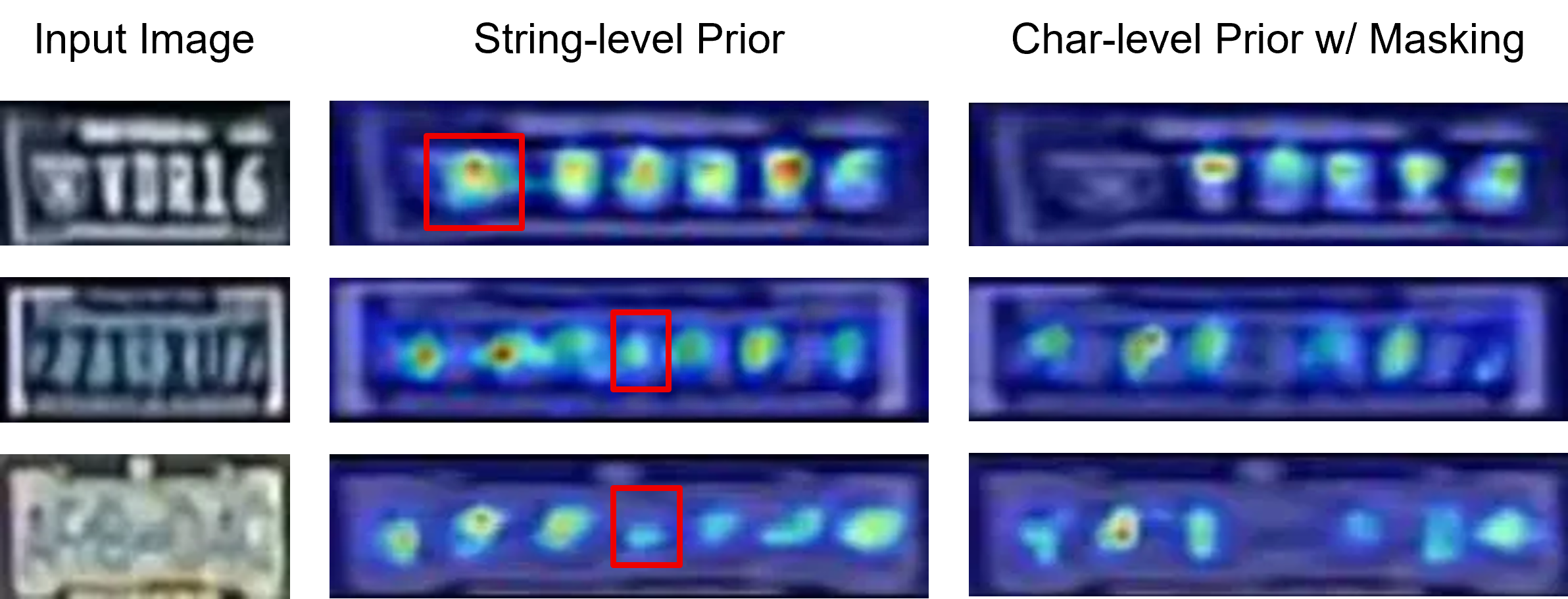}
\caption{
Cross-attention maps from the U-Net middle block comparing string-level priors and our character-level prior with CHARM. String-level priors often attend to non-character regions, whereas CHARM concentrates attention on the character masks.
}
\label{fig:attention_viz}
\end{figure}

To assess the contribution of each design choice in CharDiff-LP, we conducted ablation experiments on three components: (1) the use of text priors, (2) the granularity of guidance (string-level vs. character-level), and (3) the CHARM masking strategy.

In Tables~\ref{tab:ablation-roboflow-LP} and~\ref{tab:ablation-dashcam-LP}, the model without any text prior performs worst, indicating the importance of textual guidance. Adding a string-level prior yields a substantial gain in CER and plate-level accuracy, and replacing it with character-level priors further improves both restoration and recognition metrics, suggesting that localized guidance helps recover fine details. The full CharDiff-LP with CHARM achieves the best overall performance and generalizes best to Dashcam-LP, demonstrating that region-wise masking effectively prevents interference between characters.

For further analysis, we visualized attention maps in the U-Net middle block to compare string-level and character-level guidance (Fig.~\ref{fig:attention_viz}).
With string-level priors, attention tends to be diffuse and often spills over into neighboring characters, potentially leading to identity confusion. In contrast, CharDiff-LP yields sharp and well-localized attention that is confined to individual character regions, suggesting that CHARM effectively enforces spatially disentangled guidance and helps explain the performance gains observed in our ablation studies.

\section{Conclusion}
We presented CharDiff-LP, a diffusion-based license plate restoration model that incorporates character-level priors via the CHARM module, which applies masked cross-attention to guide each character region separately. In experiments on two datasets—Roboflow-LP with synthetic degradations and Dashcam-LP with real-world degradations—CharDiff-LP consistently outperformed recent text-guided baselines in both restoration quality and recognition accuracy, reducing CER from 11.3\% to 8.1\% on Roboflow-LP, with consistent improvements also observed on real-world Dashcam-LP images.

A current limitation is the reliance on an external segmentator and an OCR module, where severe recognition errors may lead to semantically incorrect restorations. Future work includes jointly training the recognition and restoration modules and incorporating uncertainty modeling for trustworthy deployment.

\begin{credits}
\subsubsection{\ackname} This work was supported by the National Research Foundation of Korea (NRF) grant funded by the Korea government (MSIT) (RS-2026-25482427).

\subsubsection{\discintname}
The authors have no competing interests to declare that are relevant to the content of this article.
\end{credits}
%
%
%
\bibliographystyle{splncs04}
\bibliography{references}

\end{document}